\documentclass[11pt]{article}
\usepackage[preprint]{acl}
\usepackage{times}
\usepackage{latexsym}
\usepackage[T1]{fontenc}
\usepackage[utf8]{inputenc}
\usepackage{microtype}
\usepackage{graphicx}
\usepackage{amsmath}
\usepackage{booktabs}
\usepackage{tabularx}
\usepackage{tikz}
\usetikzlibrary{arrows.meta,positioning,calc}
\usepackage{listings}
\usepackage{placeins}
\graphicspath{{figures/}}
\title{VISAGENT: An Agent-Based End-to-End Voice Interaction System}
\author{
	  \textbf{Xiang Wang\textsuperscript{1,2}},
	  \textbf{Zhaoxiang Liu\textsuperscript{1,2}},
	  \textbf{Xiang Chen\textsuperscript{1,2}},
	  \textbf{Liqiang He\textsuperscript{1,2}},
	\\
	  \textbf{Kai Wang\textsuperscript{1,2}},
	  \textbf{Shiguo Lian\textsuperscript{1,2}},
	\\
	\\
	  \textsuperscript{1}Data Science \& Artificial Intelligence Research Institute, China Unicom
	\\
	  \textsuperscript{2}Unicom Data Intelligence, China Unicom
	\\
	}
\date{}

\begin{document}
\raggedbottom
\maketitle
\begin{abstract}
Voice-based public services have evolved from manual switchboards to intelligent platforms to meet modern user needs. However, current systems remain constrained by rigid menu-driven interaction, fragmented vertical services, and limited extensibility. To address these issues, we present VISAGENT, an agent-based end-to-end voice interaction system for large-scale public services that integrates end-to-end speech interaction, hierarchical agent routing, a scalable voice-agent marketplace, and user memory support. The system replaces menu-driven workflows with natural speech and dynamically dispatches user requests to either a general conversational model or specialized domain agents via a two-level routing paradigm that achieves sub-second intent-to-agent selection over hundreds of agents. User memory informs preference-aware agent selection and supplies relevant preferences and context to selected agents, supporting personalized task execution and reducing the need for users to repeat previously provided information. A voice-agent registration mechanism converts task agents into conversational entities, enabling dynamic onboarding and ecosystem-level extensibility. We validate VISAGENT through a real-world deployment in China’s 114 hotline across multiple service domains. Experimental results demonstrate low-latency full-duplex interaction, robust routing scalability, and competitive speech naturalness. More broadly, this work advocates a shift from isolated voice assistants toward agent-centric conversational infrastructures, providing a general blueprint for building extensible voice-agent systems in public services and beyond.
\end{abstract}

\section{Introduction}
\label{sec:introduction}

Public-service hotlines provide a common entry point for users seeking information and assistance across everyday service domains. In settings such as China's 114 hotline, users may need help with transportation, accommodation, healthcare, or local services without navigating separate applications. Telephone access is particularly useful when a spoken exchange is more convenient than a graphical interface. However, a single access number does not by itself unify the services behind it. Menu navigation, repeated transfers, and separate business systems can still require users to understand how services are organized before reaching the appropriate provider.

Speech interaction models and language-model-based agents provide building blocks for a different arrangement. A conversational model can handle open-ended requests, while a domain agent can expose the tools and business processes needed to complete a particular service. Combining them requires a framework that determines when conversation is sufficient, identifies the appropriate service domain, and selects an agent from the available providers. The framework must also carry information across the handoff so that the selected agent can continue the interaction rather than asking the user to start again.

VISAGENT combines end-to-end speech interaction with hierarchical routing to provide conversational access to public services. The speech interaction model supports low-latency, full-duplex spoken communication and directly handles general conversation and question answering. A lightweight first-level router directs requests either to this conversational model or to an appropriate service domain. For domain-specific requests, an LLM-based second-level selector identifies a suitable agent by matching the user's requirements against the capabilities of agents registered within that domain. This organization connects general conversational assistance with specialized service access through a unified voice interface.

An extensible voice-agent marketplace and shared support services provide the basis for integrating heterogeneous agents into this architecture. Through a common registration mechanism, providers specify agent descriptions and callable interfaces, making their services available for discovery and invocation. Configurable voices enable agents with text-based business interfaces to deliver spoken responses suited to their roles and service contexts. A shared user memory service supplies relevant preferences and recent interaction context to both the second-level selector and the selected agent, supporting agent selection and subsequent task execution. Individual agents remain responsible for collecting task-specific information and carrying out their own business operations.

\begin{figure*}[htb]
	\centering
	\includegraphics[width=0.8\textwidth]{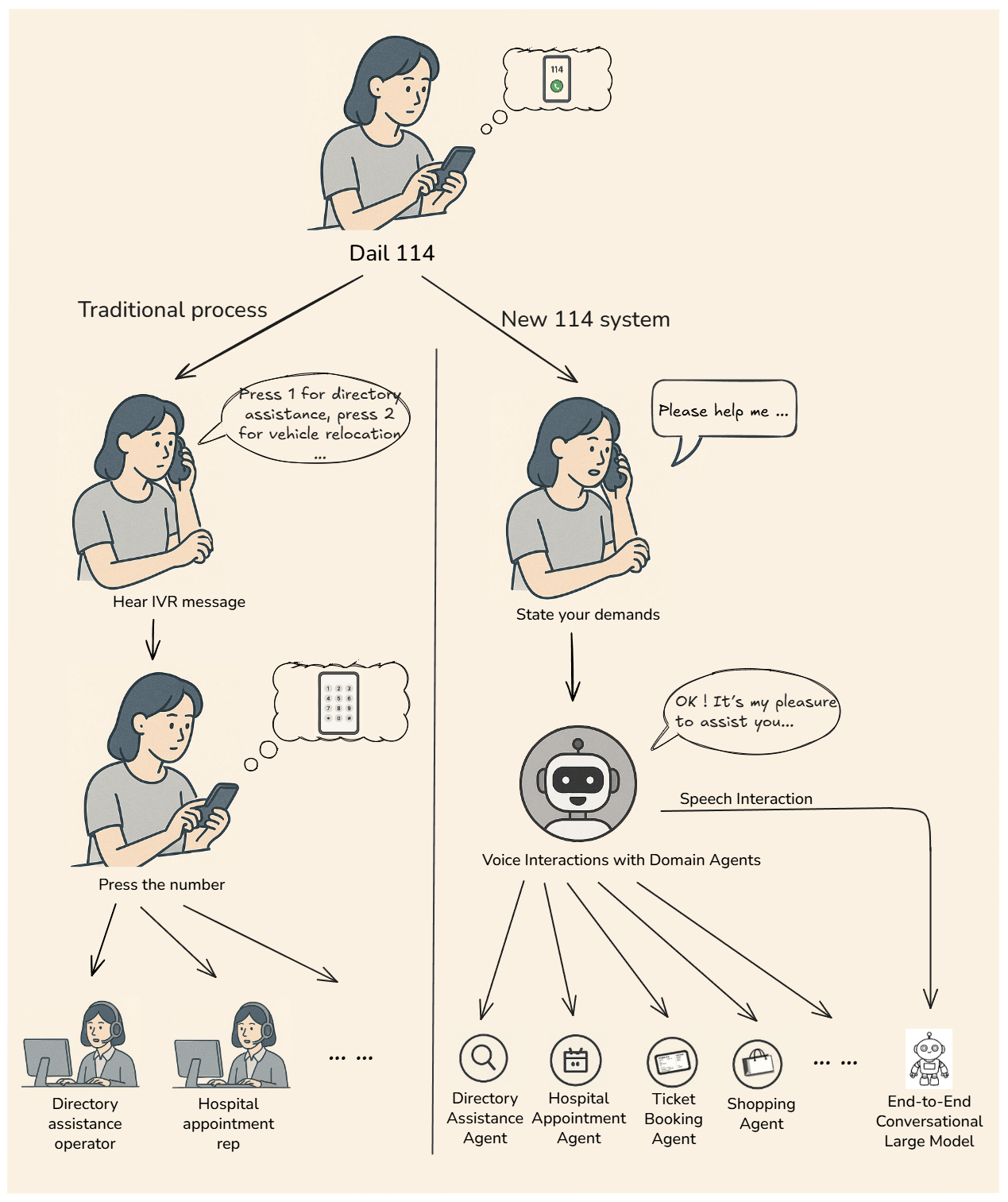}
	\caption{Interaction paradigms in the 114 hotline: traditional services rely on recorded prompts, keypad navigation, and human operators, whereas VISAGENT enables callers to express their needs directly and access domain agents or general conversational assistance through natural speech.}
	\label{fig:fig-1}
\end{figure*}

The deployment of VISAGENT in China's 114 hotline illustrates a transition from menu-driven service access to intent-driven conversational interaction, as shown in Figure~\ref{fig:fig-1}. Traditional 114 services require callers to follow recorded prompts and navigate keypad menus before being connected to human operators. VISAGENT enables callers to express their requests directly in natural language, which are then routed to domain-specific agents for service execution or to the end-to-end conversational model for general inquiries. This transition preserves the telephone-based access channel while replacing predefined menu navigation with conversational access to services.

The principal contribution of this work is an agent-centric conversational service architecture that reframes public-service hotlines from menu-driven access channels into extensible voice-agent infrastructures. By establishing a unified framework for conversational access, service discovery, agent integration, and shared user context, VISAGENT provides an architectural blueprint for integrating heterogeneous service agents into a common voice-based public-service infrastructure. The remainder of this report is organized as follows. Section~\ref{sec:related-work} reviews the research relevant to the design of VISAGENT. Sections~\ref{sec:system-overview} and~\ref{sec:components} present the overall architecture and its implementation, respectively. Section~\ref{sec:evaluation} evaluates the speech interaction and routing infrastructure. Section~\ref{sec:discussion} discusses the scope, limitations, and future development of the system, and Section~\ref{sec:conclusion} concludes the report. Illustrative examples are provided in Appendix~\ref{sec:examples}.

\section{Related Work}
\label{sec:related-work}

\subsection{End-to-End Speech Interaction}

Conventional voice-based service systems typically adopt a cascaded architecture in which automatic speech recognition, language understanding, dialogue management, and text-to-speech synthesis are implemented as separate components. Although such pipelines provide modularity and facilitate component-level optimization, errors introduced at an early stage may propagate through the remaining modules, while the sequential execution of multiple components can increase response latency and disrupt conversational continuity. Recent speech-language models have sought to reduce these limitations by incorporating speech perception, language reasoning, and speech generation within more closely integrated architectures. This development has enabled spoken dialogue systems to process acoustic input directly, generate spoken responses incrementally, and preserve information that may be lost when speech is reduced to text alone.

Representative systems illustrate several directions in this line of research. SpeechGPT connects a large language model with discrete speech representations to support cross-modal instruction following and spoken interaction \citep{zhang2023speechgpt}. Mini-Omni further investigates real-time speech generation in a compact multimodal language model \citep{xie2024mini}, while Moshi introduces a full-duplex spoken dialogue framework capable of processing and generating speech streams concurrently \citep{defossez2024moshi}. MinMo extends these capabilities through a multimodal architecture designed for seamless voice interaction \citep{chen2025minmo}. Collectively, these studies demonstrate substantial progress in speech understanding, streaming response generation, interruption handling, and conversational naturalness. Their primary emphasis, however, remains on the capabilities and evaluation of the conversational model itself. The integration of such models with large collections of independently developed service agents introduces additional architectural requirements, including intent-based dispatch, service discovery, interaction handoff, and the preservation of conversational context across system components.

Neural speech synthesis constitutes another important component of voice-based interaction. Autoregressive and flow-based synthesis models have improved speech intelligibility, prosody, speaker similarity, and streaming generation. CosyVoice~2, for example, combines language-model-based semantic representation with conditional flow matching to support scalable speech synthesis and voice generation \citep{du2024cosyvoice}, while HiFi-GAN provides an efficient neural vocoder for reconstructing high-fidelity waveforms from acoustic representations \citep{kong2020hifi}. These methods also make it possible to associate service agents with predefined or customized voices. Within VISAGENT, speech synthesis and voice cloning serve as interface technologies that allow heterogeneous agents to participate in a common spoken interaction environment.

\subsection{LLM-Based Agents and Service Routing}

The emergence of large language models has substantially expanded the capabilities of intelligent agents. Earlier task-oriented agents were generally constructed around predefined rules, domain-specific policies, or fixed action spaces. LLM-based agents can instead interpret open-ended instructions, formulate intermediate decisions, invoke external tools, and adapt their behavior according to execution feedback. ReAct established an influential formulation in which language-model reasoning is interleaved with actions and observations \citep{yao2022react}. Subsequent research has examined tool learning, planning, and task decomposition as fundamental capabilities for agents operating in external environments \citep{qin2024tool,huang2024understanding,liu2024tool}. Surveys of LLM-based autonomous agents further characterize the growing range of architectures for reasoning, memory, action, and environmental interaction \citep{wang2024survey}.

Research on multi-agent systems extends this perspective by assigning distinct roles or capabilities to multiple agents. Frameworks such as AutoGen coordinate interactions among specialized agents through structured conversational exchanges \citep{wu2024autogen}. Such systems commonly focus on collaboration among agents assembled for a particular task, including planning, execution, verification, and refinement. A service platform presents a related but distinct problem. Its agents may be developed and operated by independent providers, expose heterogeneous interfaces, and offer partially overlapping capabilities. The platform must therefore determine which agent should receive a request before the selected agent begins its own task-specific reasoning and execution. Agent selection in this setting functions as a service-discovery problem rather than as an internal step of a fixed multi-agent workflow.

\subsection{Long-Term Memory and Personalized Service Support}

Persistent memory has become an important research topic in language-model-based agents because a finite dialogue context cannot represent a user's complete interaction history. Generative Agents demonstrated how stored experiences, retrieval, and reflection can support temporally coherent behavior \citep{park2023generative}. MemGPT proposed a hierarchical memory-management architecture that coordinates information inside and outside the immediate context window \citep{packer2023memgpt}. MemoryBank further investigated the retention and retrieval of long-term conversational information for personalized interaction \citep{zhong2024memorybank}. These studies establish that agent memory involves more than retaining raw dialogue transcripts: useful memory systems must determine what information should be preserved, how it should be represented, and when it should be retrieved.

Related issues have also been studied in task-oriented dialogue. MultiWOZ provides a multi-domain benchmark for dialogue-state tracking and task completion across several service domains \citep{budzianowski2018multiwoz}, while SOLOIST investigates the construction of task-oriented dialogue systems through transfer learning and machine teaching \citep{peng2021soloist}. Such work has advanced the representation of user goals, dialogue context, database constraints, and task states. Most task-oriented dialogue systems, however, operate within a predefined ontology and a relatively fixed set of service backends. Their dialogue state primarily represents information required for the current task or session, whereas a service platform must additionally preserve reusable preferences across interactions and make them available to independently implemented agents. Taken together, existing studies primarily address either general-purpose long-term agent memory or dialogue state within predefined service ontologies, whereas VISAGENT employs compact user memory as shared contextual support for agent selection and subsequent task execution across heterogeneous service agents.

\section{System Overview}
\label{sec:system-overview}

\begin{figure*}[!t]
\centering
\includegraphics[width=0.99\textwidth]{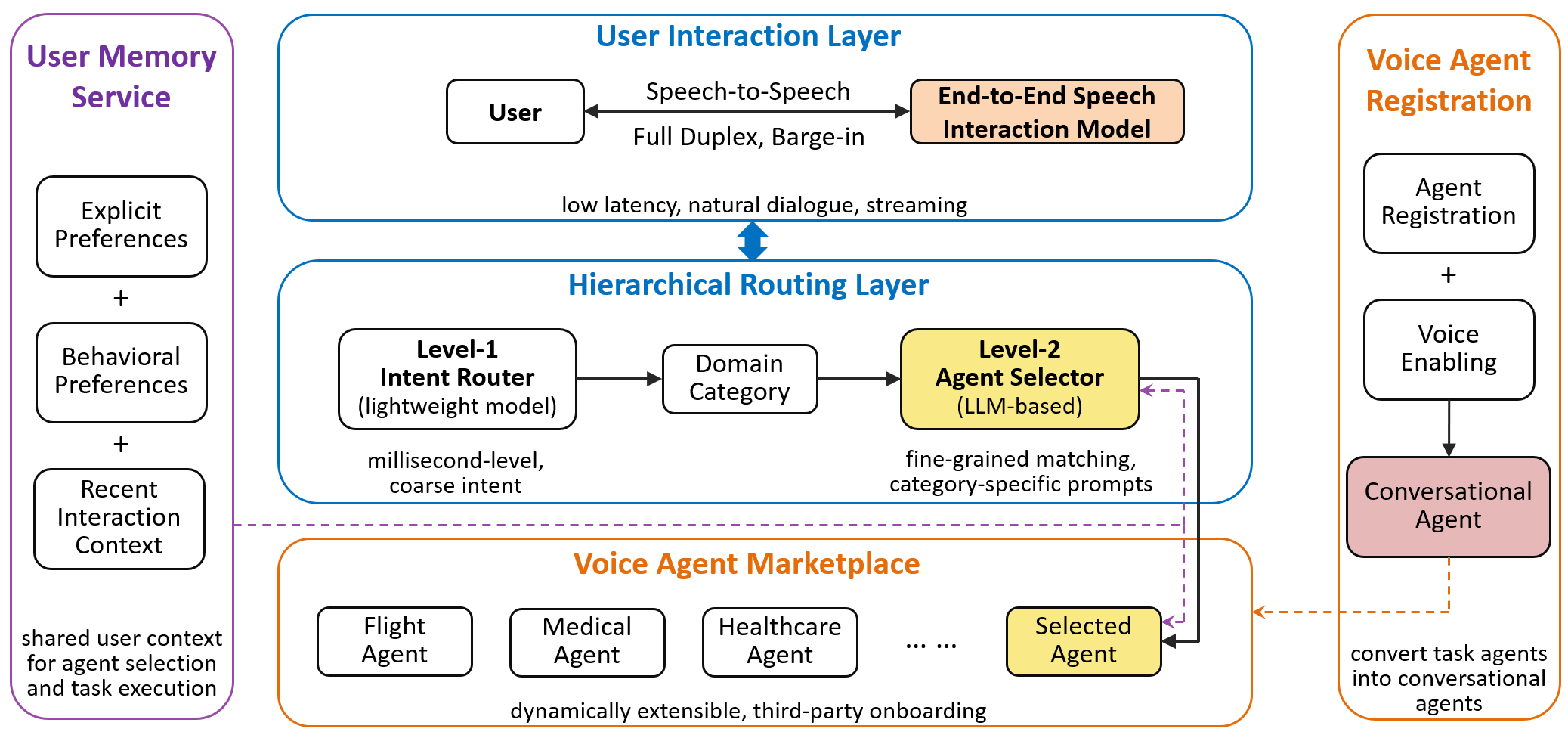}
\caption{Overall architecture of VISAGENT. The framework comprises a full-duplex user interaction layer, a two-level hierarchical routing layer, and an extensible voice-agent marketplace. The user memory service provides shared context for agent selection and task execution, while the voice-agent registration mechanism converts task agents into conversational agents and onboards them into the marketplace.}
\label{fig:architecture}
\end{figure*}

Figure~\ref{fig:architecture} presents the overall architecture of VISAGENT. The user interaction layer provides a unified speech-to-speech interface through an end-to-end speech interaction model, supporting low-latency streaming dialogue, full-duplex communication, and user barge-in. Beneath this interface, the hierarchical routing layer separates coarse intent routing from fine-grained agent selection. A lightweight Level-1 router assigns a service request to a domain or retains it in the conversational path, after which an LLM-based Level-2 selector identifies an appropriate agent within the selected domain. The voice-agent marketplace organizes agents developed by the platform and third-party providers and exposes them to the routing layer through a common registration interface. Two supporting modules extend these core layers. The user memory service maintains explicit preferences, behavioral preferences, and recent interaction context for use by the selector and the selected agent, while the voice-agent registration mechanism supplies the descriptions, callable interfaces, and voice configuration required to convert task agents into conversational agents and incorporate them into the marketplace.

\begin{figure*}[!t]
\centering
\includegraphics[width=0.8\textwidth]{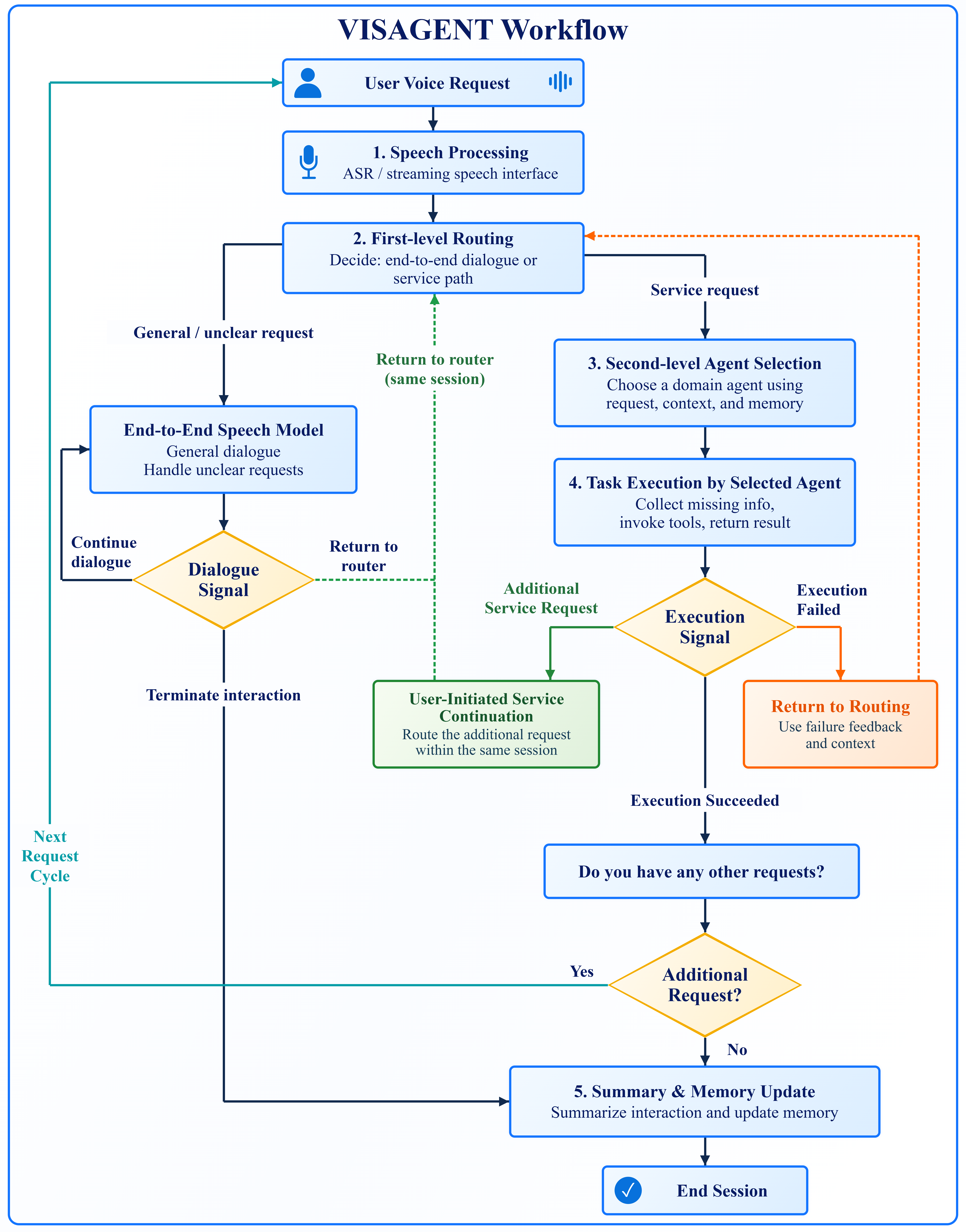}
\caption{Request-processing workflow of VISAGENT. Spoken input is routed either to the end-to-end conversational model or to a domain-specific agent selected according to the request, dialogue context, and relevant user memory. During agent execution, failures return to the first-level router with the available context and failure feedback, while additional service requests explicitly introduced by the user are redispatched within the same session. After successful task completion, the system asks whether the user has another request. An affirmative response initiates a new request cycle, whereas a negative response triggers interaction summarization, memory update, and session termination.}
\label{fig:workflow}
\end{figure*}

Figure~\ref{fig:workflow} illustrates how these components participate in request processing, from spoken interaction and intent routing to agent execution and feedback.

\paragraph{Speech interaction and intent clarification.} The speech interaction model supports direct spoken conversation, streaming responses, and full-duplex interaction with interruption handling. It also supplies intermediate transcriptions for routing and textual inputs for downstream agents. A lightweight first-level router uses the current utterance and relevant dialogue context to determine whether the request belongs to a supported service domain or should be handled by the end-to-end speech interaction model. The conversational path covers both general dialogue and requests whose service intent is not yet clear. During this path, the dialogue may continue, return to the first-level router when a service intent becomes identifiable, or terminate when no further interaction is required. When routing resumes, the accumulated dialogue context is retained so that the user does not need to restate the request. Once a service domain can be identified, the request proceeds to the second routing level. This decision requires sufficient information to identify the domain, rather than every parameter needed to complete the service.

\paragraph{Domain-specific agent selection.} The second-level router is an LLM-based agent selector operating over the agents registered in the identified domain. It uses the current request, relevant dialogue context, retrieved user memory, and registered agent descriptions to identify a suitable service agent. Restricting selection to a domain-specific candidate set reduces the amount of agent information considered for each request and separates broad service-domain identification from fine-grained agent matching. User memory provides supplementary context for distinguishing among agents with compatible capabilities, such as preferences for particular providers or service characteristics. After selection, VISAGENT passes the request, relevant dialogue context, and retrieved memory to the selected agent for task execution.

\paragraph{Agent marketplace and registration.} The voice-agent marketplace organizes services developed by the platform or by third-party providers. Registration supplies the metadata used for discovery and routing, including domain labels, functional descriptions, and callable interfaces. Providers can also associate their agents with predefined or customized voices, allowing agents with text-based interfaces to participate in the shared spoken interaction environment.

\paragraph{User memory as shared support.} The memory module maintains explicit preferences, recurring behavioral preferences, and recent interaction context derived from historical interactions. For each request, relevant records are retrieved and supplied to both the second-level selector and the selected agent. These two uses operate at different levels: a preference for a particular provider may help VISAGENT distinguish among candidate agents, whereas preferences for a hotel room with breakfast or a particular train seat are applied by the selected agent during task execution. Sharing relevant memory can reduce repeated user input and support more consistent service behavior, although the resulting action depends on how each agent interprets and applies the supplied information. Interaction summaries and service outcomes provide the basis for subsequent memory updates.

\paragraph{Task execution and feedback.} Once a service agent is selected, it manages the domain-specific workflow through the VISAGENT voice interface. The agent may collect missing information, confirm task parameters, invoke its internal tools or external services, and return the execution result. Business-specific details that are not required for first-level routing, such as a travel date, passenger information, or hotel preferences, are therefore collected during agent execution rather than by the router. During this interaction, the agent may report successful completion, execution failure, or the detection of an additional service request that falls outside its own capabilities. In the case of failure, the available dialogue context and failure feedback are returned to the first-level router. If the user introduces an additional service request, the new request and the accumulated context are likewise returned to the router for a new routing decision within the same session. After successful execution, the result is presented to the user and the system asks whether any further request remains.

\paragraph{Interaction summary and memory update.} When the conversational path terminates or the user confirms that no additional request remains after task completion, VISAGENT summarizes the interaction, records the service outcome, and updates the corresponding memory entries. Explicitly stated preferences, recurring behavioral patterns, and recent task context can therefore be incorporated into later interactions. The updated memory becomes available to subsequent routing and service execution, completing the interaction loop between speech input, routing, agent execution, feedback, and long-term personalization.

\section{System Components}
\label{sec:components}

\subsection{End-to-End Speech Interaction Model}
\label{sec:speech}

\begin{figure*}[t]
\centering
\includegraphics[width=0.6\textwidth]{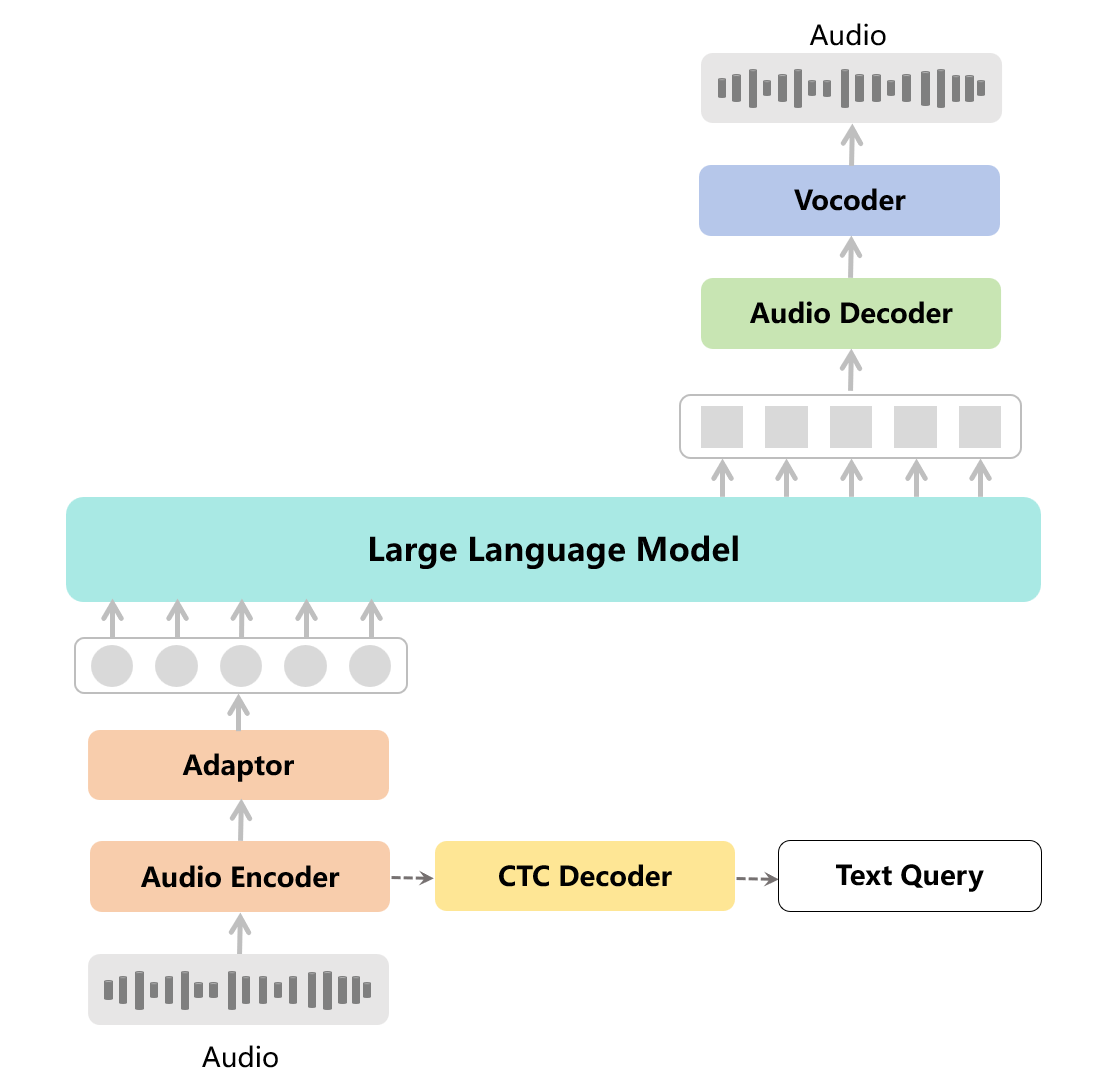}
\caption{Architecture of the speech interaction component used in VISAGENT. The primary speech-to-speech path integrates an audio encoder, an input adapter, an LLM, a streaming audio decoder, and a vocoder. An auxiliary CTC transcription branch supplies textual representations to the hierarchical routing layer and downstream service agents.}
\label{fig:speech}
\end{figure*}

The speech interaction component provides the common voice interface through which users access both general conversation and domain-specific services. As shown in Figure~\ref{fig:speech}, it consists of an audio encoder, an input adapter, an LLM, a streaming audio decoder, and a vocoder. The audio encoder extracts acoustic representations from incoming speech, and the adapter maps them into the representation space consumed by the LLM. The hidden representations generated by the LLM condition the streaming audio decoder, which produces discrete acoustic features incrementally. The vocoder subsequently reconstructs these features as an output waveform. 

In parallel with the speech-to-speech path, an auxiliary connectionist temporal classification (CTC) decoder generates intermediate transcriptions from the encoded speech representations. These transcriptions provide textual inputs for both hierarchical routing and downstream agent interaction. The routing layer uses the transcribed request together with the available dialogue context to determine an appropriate processing path, while a selected service agent receives the corresponding text and context for task-oriented interaction. VISAGENT therefore employs direct speech-to-speech generation at the user-facing interface while exposing a textual integration interface to independently implemented routing and agent components.

\begin{figure*}[t]
	\centering
	\includegraphics[width=0.8\textwidth]{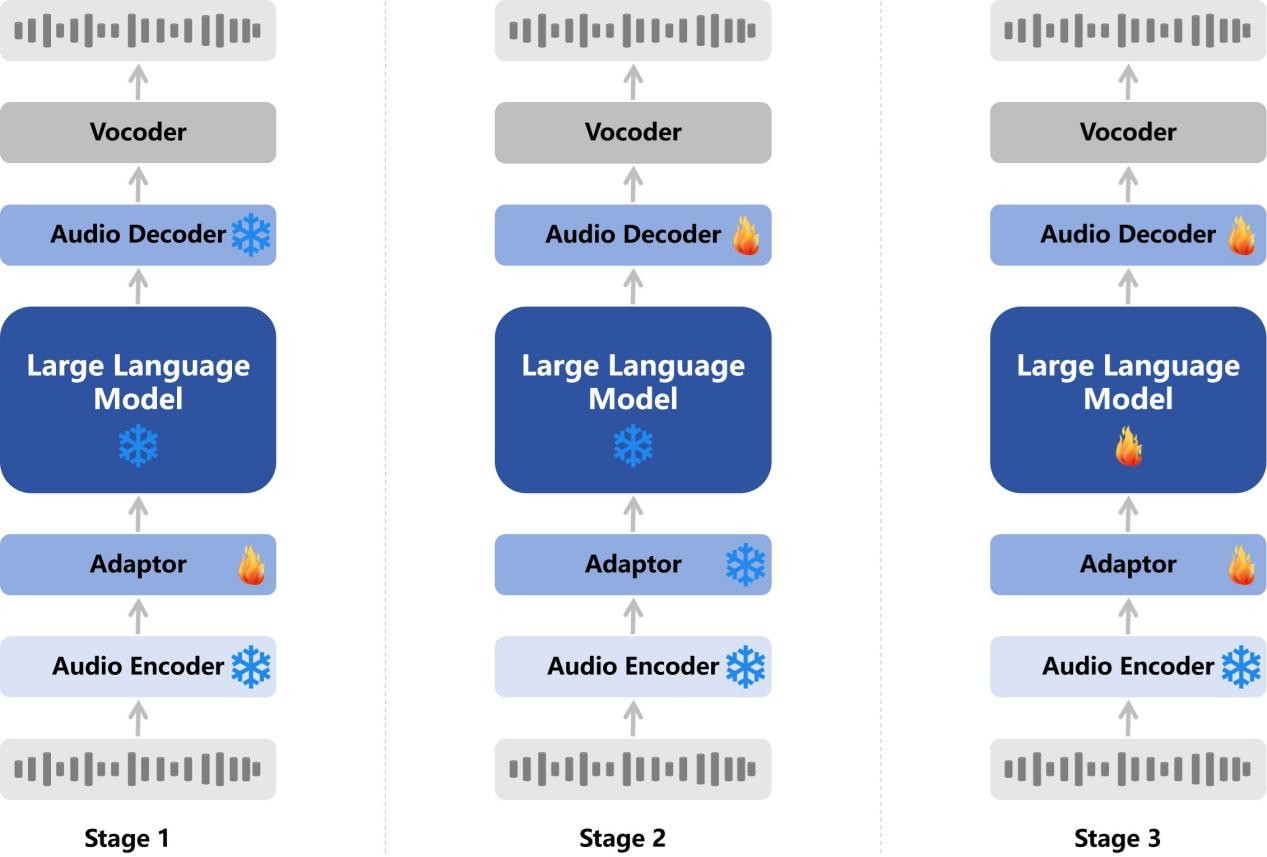}
	\caption{Three-phase training procedure for the speech interaction model. Stage~1 trains the input adapter to align the output of the frozen audio encoder with the representation space of the frozen LLM. Stage~2 trains the audio decoder to generate discrete acoustic features from the frozen LLM representations. Stage~3 jointly fine-tunes the input adapter, LLM, and audio decoder while keeping the audio encoder frozen. Snowflake and flame symbols denote frozen and trainable neural modules, respectively; the vocoder converts the decoder output into the final speech waveform.}
	\label{fig:speech-training}
\end{figure*}

\paragraph{Training configuration.}
The three-phase training procedure is illustrated in Figure~\ref{fig:speech-training}. In the first phase, the input adapter is optimized while the remaining model components are frozen, aligning the acoustic representations produced by the audio encoder with the LLM input space. In the second phase, the audio decoder is trained while the audio encoder, input adapter, and LLM remain frozen. This phase maps the hidden representations of the LLM to discrete acoustic features and incorporates streaming generation into decoder training. In the final phase, the input adapter, LLM, and audio decoder are jointly fine-tuned while the audio encoder remains frozen, allowing the speech input, language processing, and speech output components to be adapted jointly. The training corpus comprises approximately 64,000 hours of open-source Mandarin speech, 10,000 hours of open-source English speech, and 1,000 hours of private Chinese speech.

\paragraph{Full-duplex interaction.}
The implementation processes incoming speech concurrently with response generation and employs endpoint echo cancellation to support full-duplex interaction and user interruption. Users can therefore supplement, correct, or redirect an utterance while a response is being generated. Voice activity detection (VAD) identifies utterance boundaries before the resulting transcription and dialogue context are passed to the routing layer. Section~\ref{sec:latency} reports the observed interaction latency, including the VAD interval, separately from model-level latency estimates.


\subsection{Two-Level Agent Routing}
\label{sec:routing}

The hierarchical routing layer implements coarse-to-fine request dispatch over the voice-agent marketplace. Directly comparing every request with all registered agents would cause the selector's input context and inference cost to grow with the size of the marketplace. VISAGENT therefore separates routing into two decisions: a lightweight first-level router determines whether the request should remain on the conversational path or enter a supported service domain, and an LLM-based second-level selector identifies a suitable agent only among the candidates registered in that domain. This separation reduces the candidate space presented to the LLM while distinguishing domain identification from fine-grained agent matching.

\paragraph{Level 1: domain classification and conversational dispatch.}
The first-level router receives the transcribed utterance together with the available dialogue context and determines whether the request should be handled by the end-to-end speech interaction model or dispatched to a supported service domain. The current implementation uses a BERT-based intent classifier \citep{devlin2019bert}. Everyday conversation, general question answering, and other requests that do not require an external service agent remain with the speech interaction model, whereas service-oriented requests are assigned to the corresponding domain. If the service intent is not yet sufficiently clear for domain classification, the speech interaction model continues the dialogue and the request is evaluated again after clarification. The first-level router only determines the processing path and service domain; collection of task-specific parameters remains within the selected agent's workflow.

\paragraph{Level 2: memory-aware agent selection.}
Once a service domain has been identified, VISAGENT retrieves the descriptions of agents registered in that domain and the user-memory entries relevant to agent selection. The selection prompt combines the current request, pertinent dialogue context, domain-specific agent descriptions, capability and coverage constraints, and applicable user preferences. The LLM-based selector then identifies an agent whose registered capabilities satisfy the current request. User memory serves as supplementary selection context and does not override an explicit current requirement or compensate for an agent whose capabilities are incompatible with the requested service.

\subsection{Voice Agent Marketplace and Registration}
\label{sec:registration}

The voice-agent marketplace maintains a catalog of service agents organized by domain. Agents may be operated by the VISAGENT platform or by third-party providers, and multiple agents may offer overlapping services while differing in provider, geographic coverage, supported operations, and other service characteristics. The marketplace exposes their registered metadata to the routing layer without requiring access to their internal business logic, thereby separating platform-level service discovery from provider-specific task execution.

\paragraph{Registration information.}
Each registration record specifies the agent's identity and provider, service domain, natural-language description, capabilities, service tags, coverage regions, endpoint, and function specifications. The endpoint identifies the URL used to invoke a remotely hosted agent. Each function specification describes a callable operation and its corresponding input and output parameters. Together, these fields support marketplace organization, agent discovery, fine-grained selection, and service invocation. Before an agent is indexed in the marketplace, the VISAGENT platform subjects it to security review and operational testing. Only agents that pass these checks become discoverable by the routing layer and eligible for invocation. Table~\ref{tab:metadata} summarizes the roles of the registration fields.

\begin{table*}[t]
\centering\small
\begin{tabularx}{\textwidth}{@{}p{0.30\textwidth}X@{}}
\toprule
Information & Use in the framework \\
\midrule
Agent identity and provider & Identify the agent and its operator. \\
Domain, description, and capabilities & Organize candidates and describe the services available for matching. \\
Service tags and coverage regions & Describe service characteristics and geographic applicability. \\
Endpoint & Identify the URL used to invoke the selected agent. \\
Function specifications & Define callable operations and their input and output schemas. \\
\bottomrule
\end{tabularx}
\caption{Information supplied during agent registration. Appendix~\ref{sec:registration-example} gives a compact example.}
\label{tab:metadata}
\end{table*}

Once registration is complete, the agent is indexed under its designated service domain and becomes available to the hierarchical routing layer. The domain label determines the candidate group exposed to the second-level selector, while the functional description, capabilities, coverage information, and interface specification support fine-grained matching. After an agent is selected, VISAGENT invokes the registered endpoint and passes the request and relevant interaction context to the provider's implementation. The provider remains responsible for its internal task logic, external resources, and execution reliability.

\paragraph{Voice configuration and presentation.}
During registration, a provider can associate an agent with either a preset synthesis voice or a customized voice generated from a short reference recording. The current implementation provides more than 40 preset voice styles, along with speech synthesis and voice cloning services. A text-based agent can therefore receive transcribed user input and return textual responses, which VISAGENT renders through the configured voice without requiring the provider to implement a separate speech interface.

Voice profiles can be selected according to an agent's role and service context. For example, a legal-consultation agent may use a formal and serious voice, whereas an agent providing psychological counseling or emotional support may adopt a calm, gentle, and reassuring voice. Associating a stable voice profile with each agent helps users distinguish among services, improves the contextual appropriateness of spoken responses, and gives heterogeneous agents recognizable and consistent identities within the shared voice interface.

\subsection{User Memory Support}
\label{sec:memory}

The user memory service maintains compact user information that can be reused across service interactions. It provides relevant preferences and recent task context to the routing layer and selected service agents, reducing the need to repeatedly process or request information from earlier interactions. The module is responsible for memory construction, retrieval, and delivery, while the selected agent remains responsible for interpreting the supplied information within its own task workflow.

\paragraph{Representation and updates.}
VISAGENT organizes user memory into three complementary types. Explicit preference memory records preferences that users state as generally applicable, such as a preferred service provider. Behavioral preference memory captures recurring choices inferred from multiple interactions. Context memory retains recent or task-specific information, such as an active booking, a current travel plan, or the parameters of a recent service request. This separation prevents persistent preferences, recurring behavioral patterns, and temporary task context from being treated as equivalent information.

After an interaction is completed, VISAGENT constructs a compact execution summary containing the service domain, selected agent, task, invoked services, request constraints, execution details, and outcome. The summary provides the basis for updating the corresponding memory records without retaining the complete dialogue transcript as long-term memory. Appendix~\ref{sec:memory-example} illustrates these distinctions using a hotel-service example.

\paragraph{Retrieval and delivery.}
After the first-level router identifies a service domain, the memory service retrieves records relevant to the current domain and request. Structured retrieval is used for explicit and behavioral preferences, while semantic retrieval identifies related information from recent context. For each request, the memory service retrieves only the preferences, agent-use history, and recent task context relevant to the identified domain and current user need. The second-level selector and the selected agent therefore receive a compact, task-relevant memory subset rather than the user's complete interaction history.

\section{Evaluation}
\label{sec:evaluation}

This section evaluates the principal components of VISAGENT, including speech interaction performance, interaction latency, speech synthesis quality, and the accuracy and scalability of hierarchical agent routing.

\subsection{Speech Interaction Performance}
\label{sec:speech-evaluation}

We evaluate the speech interaction model on spoken dialogue tasks that assess speech understanding, reasoning, and oral response generation, and compare its performance with representative speech-enabled models. The evaluation uses URO-Bench, a speech-to-speech benchmark organized into basic and pro tracks, each containing 20 test sets at different levels of difficulty \citep{yan2025uro}. We report results on 16 open-source Chinese test sets, including nine from the basic track and seven from the pro track. Table~\ref{tab:uro} compares the 14B speech interaction model used in VISAGENT with MiMo-Audio, Step-Audio-mini, and Qwen3-Omni \citep{coreteam2025mimoaudioaudiolanguagemodels,wu2025stepaudio2technicalreport,xu2025qwen3omnitechnicalreport}.

\begin{table*}[t]
\centering\small
\begin{tabular}{llrrrr}
\toprule
Track & Test set & MiMo-Audio & Step-Audio-mini & Qwen3-Omni & VISAGENT speech \\
 & Model size & 7B & 8B & 30B & 14B \\
\midrule
Basic & Repeat-zh & 16.920 & 99.316 & 95.415 & 85.419 \\
 & LCSTS-zh & 80.280 & 81.456 & 84.369 & 82.745 \\
 & SQuAD-zh & 38.344 & 66.230 & 50.544 & 59.694 \\
 & APE-zh & 62.807 & 43.333 & 25.087 & 48.596 \\
 & MLC-zh & 22.528 & 70.804 & 32.183 & 74.942 \\
 & OpenBookQA-zh & 20.634 & 57.495 & 69.664 & 59.788 \\
 & AlpacaEval-zh & 86.802 & 75.555 & 78.639 & 78.866 \\
 & Claude-zh & 80.450 & 66.576 & 77.957 & 75.165 \\
 & Wildchat-zh & 70.546 & 77.748 & 75.473 & 78.706 \\
 & Subset average & 53.257 & 70.946 & 65.481 & \textbf{71.547} \\
\midrule
Pro & UnderEmotion-zh & 87.257 & 60.337 & 92.742 & 77.215 \\
 & CodeSwitching-zh & 64.571 & 68.380 & 72.571 & 71.619 \\
 & Safety-zh & 64.266 & 38.400 & 44.800 & 79.733 \\
 & MLCpro-zh & 30.729 & 49.479 & 13.020 & 54.166 \\
 & SRT-zh & 50.400 & 84.800 & 87.200 & 85.600 \\
 & GenEmotion-zh & 5.061 & 24.697 & 12.124 & 12.108 \\
 & GenStyle-zh & 55.213 & 84.273 & 98.461 & 85.128 \\
 & Subset average & 51.071 & 58.624 & 60.131 & \textbf{66.510} \\
\bottomrule
\end{tabular}
\caption{Results on 16 open-source Chinese test sets from URO-Bench. The subset averages are computed over the displayed test sets in each track. ``VISAGENT speech'' denotes the speech interaction model used in VISAGENT. Boldface marks the highest subset average in each track.}
\label{tab:uro}
\end{table*}

As shown in Table~\ref{tab:uro}, the VISAGENT speech interaction model achieves subset averages of 71.547 and 66.510 on the basic and pro tracks, respectively, exceeding the strongest comparison result in each subset by 0.601 and 6.379 points. Across the 16 individual test sets, it achieves first place on 4 datasets and second place on 8 datasets, underscoring its superior capabilities in advanced reasoning, safety, and open-domain conversational response generation. These results indicate that the VISAGENT speech interaction model remains strong and balanced performance across diverse spoken dialogue tasks.

\subsection{Full-Duplex Interaction Latency}
\label{sec:latency}

Full-duplex interaction latency is evaluated at both the theoretical and application levels. Table~\ref{tab:latency} reports theoretical latency estimates and the practical latency measured during system operation. The actual system delays are averaged over approximately 40 interaction rounds. VISAGENT achieves an average end-to-end latency of 1.05 seconds, which includes the 800 ms endpoint-detection interval used by the voice activity detection module.

\begin{table}[t]
\centering\small
\begin{tabular}{llr}
\toprule
Measurement & System & Latency \\
\midrule
Theoretical Latency & VISAGENT speech & 300\,ms \\
 & LLaMA-Omni & 226\,ms \\
 & MinMo & 600\,ms \\
 & Moshi & 200\,ms \\
\midrule
Practical Latency & VISAGENT & 1.05\,s \\
 & IFLYTEK SPARK & 2.32\,s \\
 & ERNIE & 2.50\,s \\
 & Doubao & 2.21\,s \\
\bottomrule
\end{tabular}
\caption{Theoretical latency and practical latency. Application-level measurements were collected on February 12, 2026.}
\label{tab:latency}
\end{table}

As shown in Table~\ref{tab:latency}, the theoretical latency of VISAGENT speech model is 300 ms, which is slightly higher than LLaMA-Omni (226 ms) and Moshi (200 ms), but substantially lower than MinMo (600 ms). In practice, however, VISAGENT achieved the lowest actual interaction latency among all tested systems, including several industry-level applications. For these deployed systems, where model architectures are undisclosed, latency was estimated from recorded interaction sessions by calculating average delays.

\subsection{Speech Synthesis Quality}
\label{sec:synthesis-evaluation}

Speech synthesis quality is evaluated on SEEDTTS test-zh by comparing the VISAGENT synthesis component with GPT-SoVITS, F5-TTS, fish-speech, and CosyVoice~2 \citep{RVC-Boss-GPT-SoVITS,chen2024f5,fish-speech-v1.4,du2024cosyvoice}. Table~\ref{tab:tts} reports Chinese character error rate (CER) for content consistency, computed with a shared ASR backend, and Normalized Mean Opinion Score (NMOS) for perceived speech quality.

\begin{table}[t]
\centering\small
\begin{tabular}{lrr}
\toprule
System & CER $\downarrow$ & NMOS $\uparrow$ \\
\midrule
GPT-SoVITS & 3.160 & 3.388 \\
F5-TTS & 1.720 & 3.474 \\
fish-speech & \textbf{1.340} & 3.489 \\
CosyVoice 2 & 1.730 & \textbf{3.500} \\
VISAGENT synthesis & 1.360 & 3.491 \\
\bottomrule
\end{tabular}
\caption{Speech synthesis results on SEEDTTS test-zh. Lower CER and higher NMOS indicate better performance.}
\label{tab:tts}
\end{table}

The VISAGENT synthesis component obtains a CER of 1.360 and an NMOS of 3.491, ranking second on both metrics among the systems in Table~\ref{tab:tts}. Its CER is close to the 1.340 obtained by fish-speech, while its NMOS is close to the 3.500 obtained by CosyVoice~2. These results demonstrate competitive content consistency and perceived speech quality on the Chinese test set.

\subsection{Routing Accuracy and Scalability}
\label{sec:routing-evaluation}

The routing evaluation uses 100 user queries and marketplace sizes of 50, 100, and 319 agents. The largest marketplace spans 15 domains, including shopping, education, psychology, law, home services, medicine, travel, government affairs, finance, culture, pets, dining, tourism, sports, and other services. Each agent is represented by a concise functional description. The single-level baseline supplies the complete uncategorized agent pool to an LLM. The two-level configuration first uses a lightweight classifier to identify the service domain and then supplies only the agents in that domain to an LLM-based selector. DeepSeek V3, Qwen3, and Kimi K2 are evaluated as selectors under both routing configurations.

\begin{table*}[t]
\centering\small
\begin{tabular}{llrrrrrr}
\toprule
 & & \multicolumn{2}{c}{50 agents} & \multicolumn{2}{c}{100 agents} & \multicolumn{2}{c}{319 agents} \\
\cmidrule(lr){3-4}\cmidrule(lr){5-6}\cmidrule(lr){7-8}
Selector & Routing & Accuracy & Latency & Accuracy & Latency & Accuracy & Latency \\
\midrule
DeepSeek V3 & Single-level & 100\% & $<1$\,s & 100\% & 4\,s & 98\% & 9\,s \\
 & Two-level & 100\% & $<1$\,s & 100\% & $<1$\,s & 100\% & $<1$\,s \\
Qwen3 & Single-level & 100\% & $<1$\,s & 100\% & 1\,s & 99\% & 8\,s \\
 & Two-level & 100\% & $<1$\,s & 100\% & $<1$\,s & 100\% & $<1$\,s \\
Kimi K2 & Single-level & 100\% & $<1$\,s & 100\% & 4\,s & 98\% & 10\,s \\
 & Two-level & 100\% & $<1$\,s & 100\% & $<1$\,s & 100\% & $<1$\,s \\
\bottomrule
\end{tabular}
\caption{Routing accuracy and latency on 100 user queries under three marketplace sizes.}
\label{tab:routing}
\end{table*}

Table~\ref{tab:routing} shows that all configurations achieve 100\% accuracy with latency below one second when the marketplace contains 50 agents. When the marketplace grows to 319 agents, single-level routing requires 8--10 seconds and achieves 98--99\% accuracy. In comparison, the two-level configuration maintains latency below one second and achieves 100\% accuracy on the evaluated query set. These results highlight the scalability and efficiency advantages of the hierarchical routing architecture for agent selection in an expanding marketplace.

\section{Discussion, Limitations, and Future Work}
\label{sec:discussion}

\paragraph{System boundaries.}
VISAGENT provides an integrated system framework for connecting users with heterogeneous service agents through a unified voice interface. It centralizes voice interaction, service discovery, routing, and memory delivery while leaving business execution with the selected provider. This separation supports extensibility, but the platform's ability to identify a suitable agent does not guarantee that the agent can complete every request. Completion also depends on available resources, tool behavior, service policies, and the agent's handling of dialogue and user context. An execution failure may return to routing, but redispatch cannot by itself remove a business constraint such as unavailable inventory.

\paragraph{Evaluation limitations.}
The routing evaluation is based on 100 user queries and marketplace configurations containing up to 319 agents, which limits the assessment of routing performance at larger scales. Future evaluations will expand both the query set and the agent pool as the marketplace grows. Practical latency is measured at the system level using commercial systems whose underlying hardware configurations cannot be controlled or standardized. The resulting measurements therefore characterize the end-to-end latency of the evaluated systems under the test conditions rather than the intrinsic latency of their speech models. The memory service has not yet been evaluated because the available interaction data remain limited. Its observed effectiveness also depends on how individual agents interpret and apply the retrieved user context, making it difficult to isolate the contribution of the memory service from downstream agent behavior. Future evaluation will therefore assess memory retrieval and memory-assisted agent selection separately from preference application and task execution, with standardized agent implementations used to control agent-specific variation.

\paragraph{Directions for future work.}
Future work will extend VISAGENT in two directions: cross-service continuation and adaptive replanning. Cross-service continuation will enable the framework to identify relationships between successive service goals, preserve task-relevant context across agent boundaries, and dispatch follow-up requests without requiring users to repeat information that has already been collected. This capability is particularly relevant to compound service scenarios, such as requesting hotel accommodation after completing a flight reservation. Adaptive replanning will allow the framework to revise an ongoing service plan when execution fails, resource conditions change, or the user modifies the original requirements. The system could then identify feasible alternatives, reconstruct the required sequence of agent calls, and obtain user confirmation before continuing execution. Future research will investigate the context-transfer mechanisms, plan-validation methods, and recovery policies required to implement and evaluate these capabilities reliably.

\section{Conclusion}
\label{sec:conclusion}

This paper presented VISAGENT, an integrated system framework for voice-based access to heterogeneous service agents. VISAGENT combines end-to-end speech interaction, hierarchical agent routing, voice-agent registration, an extensible agent marketplace, and user memory support within a unified architecture. Its two-level routing design separates efficient domain identification from fine-grained agent selection, while the registration mechanism provides a standardized means of incorporating third-party task agents into the voice service ecosystem. User memory further supplies task-relevant context to both agent selection and service execution. This organization separates reusable platform capabilities from provider-specific business logic and supports the continued expansion of voice-based services.

The experimental results demonstrate competitive performance in spoken dialogue and speech synthesis, low-latency full-duplex interaction, and the scalability and efficiency advantages of hierarchical routing across the evaluated marketplace configurations. The deployment of VISAGENT in China's 114 hotline further demonstrates the feasibility of the framework in a real-world public-service environment. These results establish VISAGENT as a practical architectural foundation for building extensible voice-agent service systems.

\FloatBarrier
\bibliography{references/report}
\clearpage
\appendix
\section{Illustrative Examples}
\label{sec:examples}

This appendix provides illustrative examples of agent registration, user-memory construction, and agent-mediated service interaction. The three examples address different system interfaces and do not form a single execution trace. They are shortened for readability and are not presented as experimental observations.

\subsection{Agent Registration Example}
\label{sec:registration-example}

The courier example describes an agent for delivery orders, pickup booking, and package tracking. Its metadata card contains the information used for marketplace discovery and selection, together with the function specifications required to invoke the agent after handoff. The provider name is anonymized, and the record is presented as an illustrative registration example rather than evidence of a verified production integration.

\begin{lstlisting}
{
  "agent_id": "courier_sf_01",
  "agent_name": "XX Express Courier Agent",
  "provider": {
    "name": "XX Express",
    "type": "third_party"
  },
  "domain": "courier",
  "description": "A courier agent for fast parcel delivery, pickup booking, and package tracking.",
  "capabilities": [
    "package_delivery", "pickup_booking",
    "package_tracking"
  ],
  "service_tags": [
    "fast_delivery", "high_reliability",
    "door_to_door_pickup"
  ],
  "coverage_regions": [
    "Beijing", "Shanghai",
    "Guangzhou", "Shenzhen"
  ],
  "endpoint": "https://api.example.com/courier",
  "functions": [
    {
      "name": "order_delivery",
      "inputs": [
        {
          "name": "sender_address",
          "type": "string",
          "description": "pickup address"
        },
        {
          "name": "receiver_address",
          "type": "string",
          "description": "delivery address"
        },
        {
          "name": "pickup_time",
          "type": "string",
          "description": "preferred pickup time"
        }
      ],
      "outputs": [
        {
          "name": "order_id",
          "type": "string",
          "description": "created delivery order identifier"
        },
        {
          "name": "estimated_delivery_time",
          "type": "string",
          "description": "estimated delivery time"
        }
      ]
    },
    {
      "name": "track_package",
      "inputs": [
        {
          "name": "tracking_number",
          "type": "string",
          "description": "package tracking number"
        }
      ],
      "outputs": [
        {
          "name": "package_status",
          "type": "string",
          "description": "current package delivery status"
        }
      ]
    },
    {
      "...": "..."
    }
  ]
}
\end{lstlisting}

The domain, capabilities, service tags, and coverage regions support marketplace retrieval and agent selection. The endpoint identifies the URL used to invoke the selected agent, while the function definitions provide the input and output schemas needed to construct an executable request. The router uses these descriptions for discovery and handoff but does not execute the courier workflow itself.

\subsection{User Memory Example}
\label{sec:memory-example}

The hotel example illustrates how a historical interaction is converted into a compact execution summary and subsequently contributes to structured user memory. The interaction begins with a request for a one-night stay in Suzhou on October~4. The user then provides a budget, room preference, breakfast requirement, and preferred location before selecting one of the options returned by the agent.

\begin{lstlisting}
{
  "interaction_id": 2,
  "domain": "hotel",
  "selected_agent": {
    "agent_id": "hotel_xx_01",
    "agent_name": "XX Hotel Booking Agent",
    "provider": "XX Travel"
  },
  "messages": [
    {
      "role": "user",
      "content": "Help me find a hotel in Suzhou for one night on Oct. 4."
    },
    {
      "role": "agent",
      "content": "Please provide budget, room type, and breakfast preference."
    },
    {
      "role": "user",
      "content": "Within 600 RMB, king-bed room, preferably with breakfast, near Suzhou Industrial Park."
    },
    {
      "role": "agent",
      "content": "Option 1: Suzhou Industrial Park Business Hotel, king-bed room with breakfast, 430 RMB/night. Option 2: ..."
    },
    {
      "role": "user",
      "content": "Book the first hotel."
    },
    {
      "role": "agent",
      "content": "Please provide the information of the occupants."
    },
    {
    	"...": "..."
    }
  ]
}
\end{lstlisting}

The interaction is converted into the following execution summary:

\begin{lstlisting}
{
  "user_id": 1,
  "interaction_id": 2,
  "domain": "hotel",
  "task": "hotel query/search and booking",
  "operation": "search_and_book",
  "selected_agent": {
    "agent_id": "hotel_xx_01",
    "agent_name": "XX Hotel Booking Agent",
    "provider": "XX Travel"
  },
  "services_used": [
    "hotel_search", "hotel_booking"
  ],
  "request_constraints": {
    "date": "Oct. 4",
    "preferred_location": "Suzhou Industrial Park",
    "preferred_room_type": "king_bed",
    "breakfast_preferred": true,
    "budget_limit": "within_600_RMB"
  },
  "booking_details": {
    "hotel_name": "Suzhou Industrial Park Business Hotel",
    "location": "Suzhou Industrial Park",
    "room_type": "king_bed",
    "breakfast_included": true,
    "price": "430 RMB/night"
  },
  "outcome": "success"
}
\end{lstlisting}

After aggregating similar execution summaries, the user's memory is updated as follows. Explicit preference memory remains empty because, up to this point, the user has not explicitly stated any personal preference that applies generally across service interactions. The entries in behavioral preference memory are derived from recurring patterns observed across multiple hotel-booking records rather than from the single interaction shown above.

\begin{lstlisting}
{
  "user_id": user_1,
  "explicit_preference_memory": {},
  "behavioral_preference_memory": {
    "accommodation": {
      "preferred_room_type": "king_bed",
      "breakfast_preferred": true,
      "typical_budget": "within_600_RMB",
      "frequently_used_agent": {
        "agent_id": "hotel_xx_01",
        "agent_name": "XX Hotel Booking Agent",
        "provider": "XX Travel"
      }
    },
    "workflow_patterns": [
      "hotel_search -> hotel_booking",
      "..."
    ]
  },
  "context_memory": {
    "recent_accommodation": {
      "domain": "hotel",
      "task": "hotel query/search and booking",
      "selected_agent": {
        "agent_id": "hotel_xx_01",
        "agent_name": "XX Hotel Booking Agent",
        "provider": "XX Travel"
      },
      "key_parameters": {
        "location": "Suzhou Industrial Park",
        "room_type": "king_bed",
        "breakfast": true,
        "budget": "430 RMB/night",
        "...": "..."
      },
      "outcome": "success"
    }
  }
}
\end{lstlisting}

This example distinguishes recurring accommodation patterns from the details of a recent booking. Behavioral memory records patterns aggregated across interactions, including an agent that has been used repeatedly, whereas context memory retains the agent, parameters, and outcome associated with the recent accommodation task. During a later hotel request, VISAGENT can supply the relevant agent-use history and provider information to the selector and pass execution-related preferences and context to the selected hotel agent.

\clearpage
\onecolumn
\subsection{Request Routing and Agent Interaction}
\label{sec:interaction-example}

\begin{figure}[!ht]
\centering
\includegraphics[width=0.87\textwidth,height=0.60\textheight,keepaspectratio]{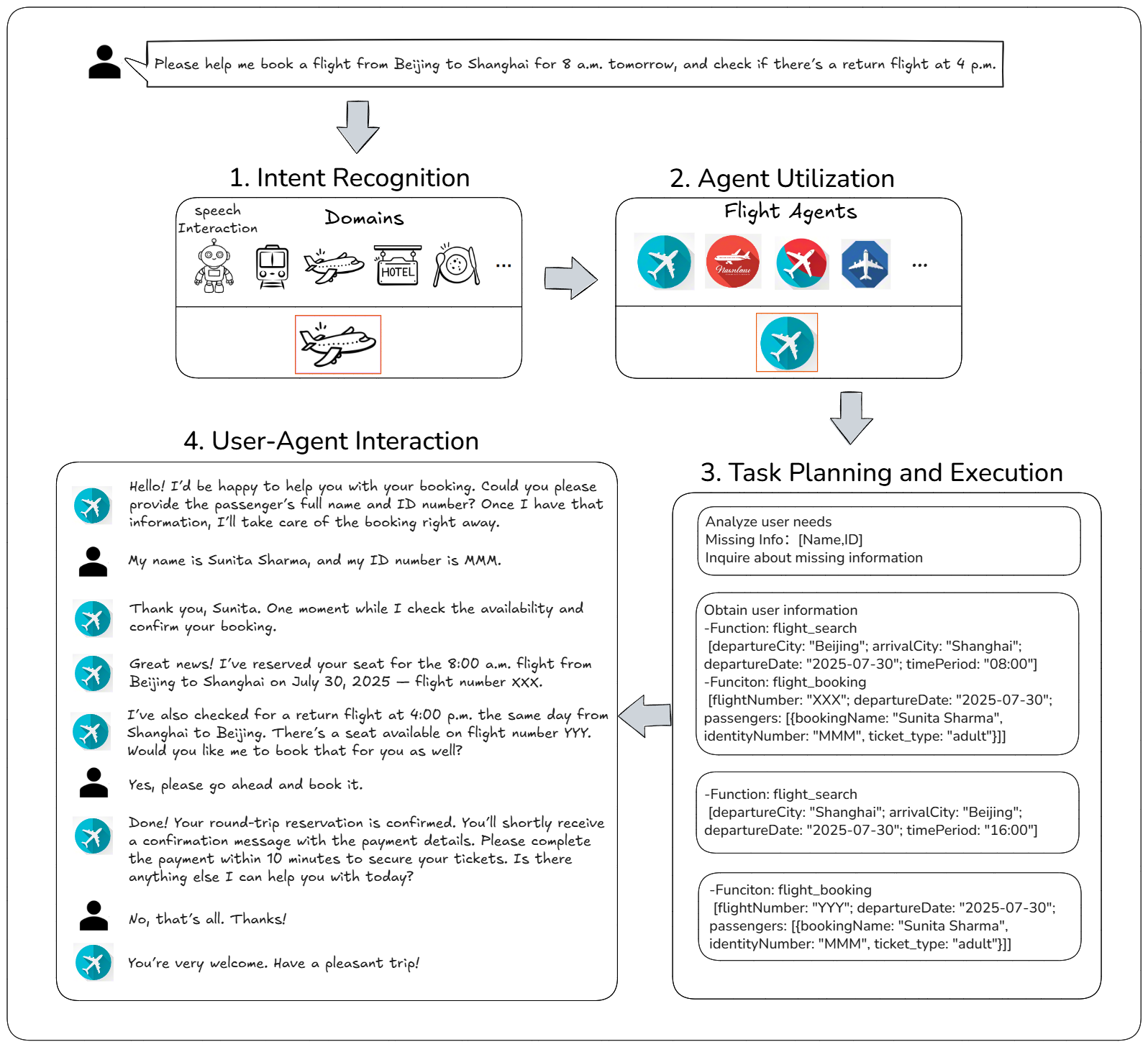}
\caption{Illustrative flight-booking interaction in VISAGENT. The hierarchical routing process identifies the flight domain and selects a flight agent. The selected agent plans the task, collects missing passenger information through spoken interaction, and invokes flight-search and booking functions to complete the outbound and return reservations.}
\label{fig:flight-example}
\end{figure}

Figure~\ref{fig:flight-example} illustrates an agent-mediated flight-booking interaction. The user requests an outbound flight from Beijing to Shanghai at 8 a.m. the following day and a return flight at 4 p.m. The first routing level identifies the flight domain, and the second-level selector chooses an appropriate flight agent. The selected agent then collects the required passenger information and invokes its search and booking functions to complete the request.

The example illustrates the division of responsibility between the VISAGENT framework and the selected service agent. VISAGENT performs domain classification and agent selection before transferring the request and available context to the flight agent. The agent determines which task parameters are missing, obtains the necessary information through dialogue, and invokes the registered functions required for execution. Because both flight directions are specified in the initial request and handled by the same agent, the interaction constitutes a single domain-specific service task.
\FloatBarrier

\end{document}